\documentclass{mem}
\usepackage{balance}
\usepackage{amsmath}
\usepackage{flushend}
\usepackage{graphicx}
\usepackage{caption}
\usepackage{algorithm}
\usepackage{algpseudocode}
\usepackage{blindtext}
\usepackage{authblk}
\usepackage[breaklinks,pdftex]{hyperref}
\begin{document}
\title{Managed Geo-Distributed Feature Store: Architecture and System Design}
\author{Anya
Li, Bhala Ranganathan\thanks{System architect and tech lead of AzureML Feature Store.}, Feng Pan\thanks{Engineering manager}, Mickey Zhang, Qianjun Xu\thanks{Engineering manager}, Runhan Li, Sethu Raman\thanks{Product Manager of AzureML Feature Store who structured and organized the product vision and specifications. }, Shail Paragbhai Shah, Vivienne Tang\\
Microsoft\thanks{All the authors are from the AzureML Feature Store product group and are listed in alphabetical order.}}

\authorrunning{ Managed Feature Store }
\titlerunning{ Architecture and System Design }
\abstract{
Companies are using machine learning to solve real-world problems and are developing hundreds to thousands of features in the process. They are building feature engineering pipelines as part of MLOps life cycle to transform data from various data sources and materialize the same for future consumption. Without feature stores, different teams across various business groups would maintain the above process independently, which can lead to conflicting and duplicated features in the system. Data scientists find it hard to search for and reuse existing features and it is painful to maintain version control. Furthermore, feature correctness violations related to online (inferencing) - offline (training) skews and data leakage are common. Although the machine learning community has extensively discussed the need for feature stores and their purpose [\cite{mj_2022, managing-ml-pipeline}], this paper aims to capture the core architectural components that make up a managed feature store and to share the design learning in building such a system.
}
\date{}
\institute{}
\journalname{}
\maketitle{}
\section{Introduction}

Features are the input data for machine learning model. For data driven use cases in an enterprise context, features are often transformations on historical data (aggregation and row-level transformations). Consider a machine learning  model for customer churn. The inputs to the model could include customer interaction data like \verb|30day_transactions_sum|, \verb|30day_complaints_sum|.

Feature store is a centralized system for storage, management, and retrieval of features. Feature store experience primarily addresses the pain points below:

    \noindent {\bf Search and reuse features:} to avoid redundant work and deliver consistent predictions. 

    \begin{quote}
        Features should be easily searchable and discoverable across different teams. The ability to browse metadata, lineage, and feature statistics to understand a feature and choose the right feature in training/inference; the ability to reuse the transformation logic that generates a feature; the ability to track features used in a model to avoid requiring manual effort to cherry-pick features.
    \end{quote}

    \noindent {\bf Manage the feature materialization feature engineering pipelines:} free the team from the operational aspects of feature engineering. 
    \begin{quote}
        Creating, monitoring, and maintaining feature engineering pipelines is non-trivial infrastructure work. Having a managed solution enables the team to focus on core feature engineering work. Materialized feature data is efficient when there are complex feature calculations that are reused across models.
    \end{quote}

    \noindent {\bf Agility of feature development:} 
    \begin{quote}
        Data scientist can use DSL or language/framework (e.g. pandas, pyspark) they are familiar with to define transformations.
        They can easily test the feature in local development. Once the feature is ready, data scientist can deploy the features to production environment without help from the data engineering team.
    \end{quote}

    \noindent {\bf Avoid offline and online data skew:}  
    \begin{quote} 
        Feature store maintains the consistency of data in offline and online data stores to avoid training/inference skew.
    \end{quote}

    \noindent {\bf Ability to retrieve data in a point-in-time correct manner:} 
    \begin{quote}
        Models should not accidentally use data in the future to prevent data leakage issues. Implementing this by hand is complex and error prone.
    \end{quote}

\section{System Overview and Core Concepts}

    \subsection{System Overview}\label{sec:overview}

    \begin{figure}[h]
    \centering
    \resizebox{\hsize}{!}{\includegraphics[clip=true]{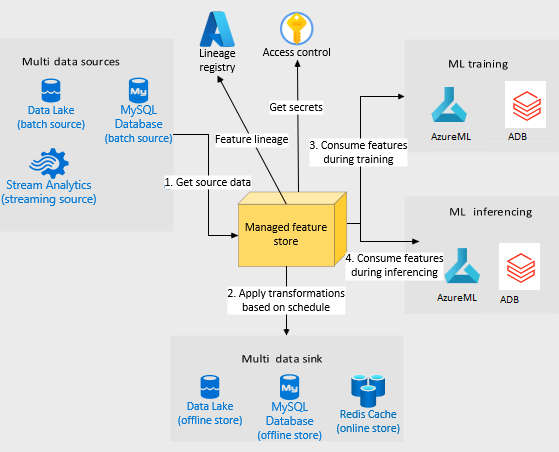}}
    \caption{\footnotesize
    System Overview
    }
    \label{fig:overview}
    \end{figure}
    
    Conceptually, a managed feature store in Figure \ref{fig:overview} provides the functionalities listed below :
    \begin{itemize}
        \item {\bf Feature store management:} Create, Delete, Search of feature stores.
        \item {\bf Feature store asset management:} Create, Delete, Update, Search of assets that belong to a given feature store.
        \item {\bf Feature governance:} RBAC, Compliance
        \item {\bf Feature engineering:}
        \begin{enumerate}
            \item Recurrent feature materialization based on a schedule.
            \item On-demand backfill materialization.
            \item Offline feature retrieval to support point-in-time joins with high data throughput.
            \item Online feature retrieval to support feature retrieval with low latency.
            \item Monitor feature usage and lineage.
        \end{enumerate}
        \item {\bf Feature store execution modes:}
        \begin{enumerate}
            \item Bring your own - Support customer-provisioned online/offline stores.
            \item Managed - Managed offline/online stores for better SLAs.
            \item One box – A local development experience before registering for managed feature store.
        \end{enumerate}
        \item {\bf Regional presence:} Available in many regions and provides geo-distributed access to feature store assets in training and inference.
        \item {\bf Enterprise grade SLAs:}
        \begin{enumerate}
            \item Primitive guarantees such as availability, consistency, latency, fail over etc.
            \item Data Staleness/Freshness: This metric indicates how fresh or latest is the feature data computed by the platform.
        \end{enumerate}
    \end{itemize}

    \subsection{Conceptual overview}\label{sec:conceptualoverview}
    Feature sets are a collection of features that are generated by applying transformation on a source system data. Feature sets encapsulate a source, the transformation function, and other parameters relevant for the transformation logic. It also additionally defines managed capabilities such as materialization settings. \\    
    
    \noindent Entities define index/key columns for feature lookup and join. Typically, entities are created once and reused across feature sets. Besides representing the index columns, it also serves the purpose to organize the feature sets. \\
   
    \noindent The transform function defined in feature sets is expected to output a tabular data (e.g. dataframe) which includes the following in its schema: 1) index columns, 2) timestamp column and 3) all the feature columns defined by the in feature set.

\section{System architecture}
    \begin{figure}[t]
    \centering
    \resizebox{\hsize}{!}{\includegraphics[clip=true]{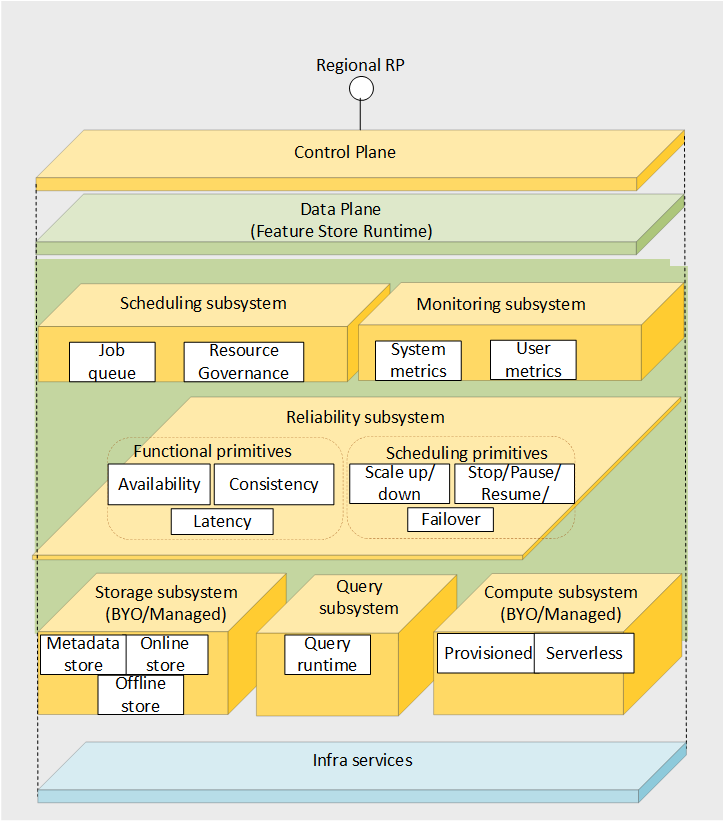}}
    \caption{
    \footnotesize
    System architecture}
    \label{fig:arch}
    \end{figure}

    In this section, we talk about the system architecture based on the design of AzureML Feature Store. Certain design aspects described in this section and section \ref{sec:mechanism} remain in the aspirational phase, and have not been formally implemented in the AzureML Feature Store at the time this paper is published. These components represent our strategic ideation and offer a glimpse into the prospective capabilities of the product. For information regarding currently available functions and features in AzureML Feature Store, please refer to the official documentations which can be accessed via the following link: https://aka.ms/featurestore-get-started.
    
    \subsection{Anatomy of a managed feature store}\label{sec:anatomy}
    Figure \ref{fig:arch} shows the overall system architecture, and we discuss the key subsystems and mechanisms in the below sections.
        \subsubsection{Context aware scheduling}
        The scheduling subsystem is responsible for ensuring the following.
        \begin{itemize}
            \item Efficient and cost-effective usage of compute capacity.
            \item Backfill materialization should not conflict with system scheduled materialization. i.e., a context aware partitioning scheme is used intelligently to define the distribution or coalescing of feature windows used in each unit of feature computation. In one implementation such a partitioning scheme can be obtained from customers optionally.
            \item When the customer performs a backfill materialization, we may want to temporarily suspend existing scheduled system materializations and resume later.
        \end{itemize}
        \subsubsection{Improved fault tolerance}
        \begin{itemize}
            \item A health/monitoring subsystem provides rich insight into the health of the system. We classify these broadly as built in (system) and custom (user defined) metrics. The metrics visible to customers could provide useful insights into their feature engineering process and further fine tune their workloads. The system metrics are used to ensure high availability and other enterprise promises.
            \item When one region is down, we may want to use the resources from cross regions to ensure high availability. Also, when the runtime comes back up, we need to make sure it can safely resume from where it left off without any data loss.
        \end{itemize}

        \subsubsection{Improved reliability}
        \begin{itemize}
            \item The system should monitor action status, retry failed actions, and create alerts for non-recoverable failures. The system should overall have minimal downtime aligning to the SLA guarantees. 
            \item There should be consistent results served between online and offline stores with no data loss or data leaks.
            \item We want to scale up or down the managed resources like Redis to meet the HA and throughput requirements of customers when accessing the feature store.
            \item The latency for feature retrieval should be reasonable without additional overhead especially for cross region use cases. In one implementation, the assets could be geo replicated. In other implementations, it could be cross region access instead of geo-replicating assets due to data governance compliance requirements.
        \end{itemize}
        
        \subsubsection{Storage subsystem}
        \begin{itemize}
            \item Metadata store persists information about feature store assets (static content) and system runtime state. 
            \item Online stores (data sink) provide low latency retrieval guarantees. e.g., Redis [\cite{redis_2023}].
            \item Offline stores (data sink) provide big data processing, low cost, high throughput retrieval guarantees e.g., Azure Data Lake Storage Gen2 [\cite{delta-lake}].
        \end{itemize}

        \subsubsection{Compute subsystem}
        The system may use a pre-provisioned managed compute or accept bring your own compute. The managed compute uses Apache Spark for large-scale feature calculation. The configured compute has a serverless architecture for a cost-effective offering.
        
        \subsubsection{Optimized query execution}
        A query engine using DSL [\cite{feathr-ai}] may be used to optimize feature transformations for optimized feature computation. When customers define features using UDF, feature store treats the UDF as a black box and it depends on compute to optimize the query plan. However, when customers define features using DSL (a common case is rolling window aggregation), feature store can optimize the aggregation based on join results to reduce the compute cost.
        
        \subsubsection{Python SDK}
        The feature store SDK for Python enables customer to write code to manage and interact with feature store resources. The SDK leverages PySpark to perform feature calculations, data materialization, and other feature set functionalities. These details are discussed in depth in Section \ref{sec:mechanism}.
        
    \subsection{Resource model}\label{sec:rm}
    A feature store is a separate RESTful resource and globally accessible. A feature store contains several artifacts and assets that define the concepts within the feature store. These concepts may differ with different implementations but in general such assets are usually required to manage feature engineering pipelines. Each feature store also comes with a materialization policy and other operational policies. There may be one or many feature engineering pipelines scheduled based on the configured policy.

    \begin{figure}[h]
    \centering
    \resizebox{\hsize}{!}{\includegraphics[clip=true]{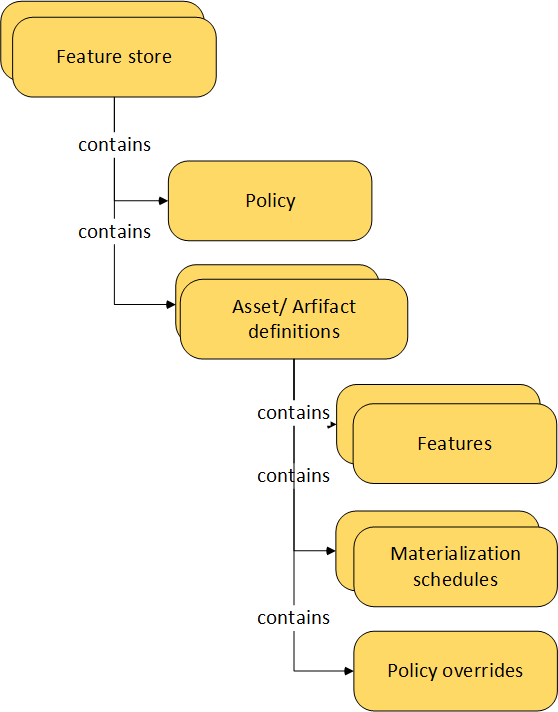}}
    \caption{
    \footnotesize
    Resource model}
    \label{fig:rm}
    \end{figure}

\section{Key mechanisms}\label{sec:mechanism}
    \subsection{Asset metadata management and versioning}
    Feature store contains several assets like feature sets and feature entities. Such assets are versioned because each such asset properties could either be mutable or immutable. When an asset’s immutable property needs to be modified, we expect its version to be incremented instead. For example, there can be several versions of a feature set and each such version could have its own materialization schedule. The defined transformation code could be unique per feature set version and hence it could be categorized as an immutable property.
    
        \subsubsection{Hub and spoke architecture}
        In addition to asset versioning and source control, it is also important to enable asset sharing and reuse across teams. We designed a hub and spoke model where feature store is the hub, and the consuming machine learning is the spoke. This architecture allows us to use assets across subscriptions. Such a model helps us avoid customers to model feature stores using peer-to-peer architecture which only allows the same feature store to be the consuming workspace. 
        
        \subsubsection{Asset access within and cross regions}
        Feature stores can be created in any of the supported regions and several assets belong to a given feature store. There are two mechanisms with which we could make an asset belonging to one feature store accessible to a consuming workspace in a different region. One such mechanism is to geo-replicate the assets for improved latency, especially for online store retrieval because data is available locally, but such an approach may not be possible in geo-fenced architectures due to data compliance issues. Another approach is to use access control to provide cross-region access to specific assets. In this case data resides in the location where it was created. Our current implementation uses the latter approach but we also have the former approach in the road map to suit more specific use cases where the latter approach is not suitable.

        \begin{figure}[h]
        \centering
        \resizebox{\hsize}{!}{\includegraphics[clip=true]{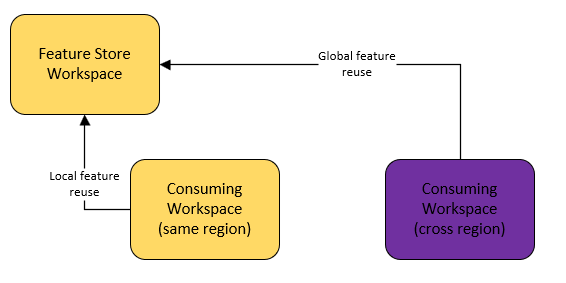}}
        \caption{
        \footnotesize
        Assets sharing within and cross regions}
        \label{fig:hubspoke}
        \end{figure}
        
    \subsection{Feature calculation}
    Feature values are calculated by applying the feature transformation logic on top of the source. The transformation logic is defined by a user-defined function (UDF) \(udf(source\_df, context) \to feature\_df\) in the code folder. While the customer can define the UDF as they want, we require the output \verb|feature_df| to have:
    \begin{itemize}
        \item Index columns
        \item Timestamp column (if feature set defines timestamp column)
        \item All the feature columns as defined by the schema in the feature set
    \end{itemize}

    The same calculation flow is applied when:
    \begin{itemize}
        \item Materialize the feature set (incremental or one-time backfill)
        \item Offline join of multiple feature sets when the feature sets are not materialized.
    \end{itemize}

    Snippet \ref{alg:fscalc} provides intuition on how feature calculation works.

    \begin{algorithm*}
    \caption{Feature calculation snippet}\label{alg:fscalc}
    \begin{algorithmic}
    \Require
    \\ 
    $feature\_window\_start\_ts > 0$\\
    $feature\_window\_end\_ts > 0$\\
    $source\_lookback \geq 0$\\
    \\
    /*Define the source data time window according to feature window*/
    \State $source\_window\_start\_ts \gets feature\_window\_start\_ts - 
    source\_lookback$
    \State $source\_window\_end\_ts \gets feature\_window\_end\_ts$
    \\
    \State /*Read source table into a dataframe*/

    \State $df1 = spark.read.parquet(source.path).filter(df1["ts"] \ge source\_window\_start\_ts \land df1["ts"] < source\_window\_end\_ts)$
    \\
    \State /*Apply the feature transformer*/\\
    $df2 = FeatureTransformer.\_transform(df1)$
    \\
    \\
    /*Filter the feature(set) to include only feature records within the feature window*/
    
    \State $feature\_set\_df = df2.filter(df2["ts"] \ge feature\_window\_start\_ts \land df2["ts"] < feature\_window\_end\_ts)$
    
    \end{algorithmic}
    \end{algorithm*}

    \subsection{Feature Materialization}
    
    Complex transformation/aggregation logic can be expensive to compute (time/money). By materializing it, training and inference jobs can run faster since they can now use pre-computed feature values. In the case of online inference, materialization can help with low latency serving and help avoid training/serving skew.
        
    Materialization jobs will compute feature values along the timeline and store them in the offline/online stores. The jobs use the same feature calculation logic as in the section before.
    
    There are two types of materialization jobs:
    \begin{itemize}
        \item One-time backfill: job triggered as demanded by user that covers one feature window defined by user
        \item Scheduled incremental materialization: job triggered by a system schedule using the cadence defined by user. Each job covers an incremental window on the feature event timeline.
    \end{itemize}

    The scheduling subsystem in feature store should track multiple states regarding the materialization:
    \begin{itemize}
        \item Data state: along the feature event timeline, the state of feature data materialization of any given time window. The basic states can be not-materialized and materialized. 
        \item Job state: tracking active materialization jobs (including queued and running ones) and the feature window each job covers.
    \end{itemize}

    By tracking the states, the scheduling subsystem will make sure that:
    \begin{itemize}
        \item Concurrent jobs do not have overlapping feature windows. This ensures that the feature data in the offline/online store do not have nondeterministic results when multiple jobs update the data at the same time.
        \item When feature retrieval (offline and online) returns no result, there is clear distinction between: 1) feature data is not materialized in the requested feature window; 2) there is no feature data in the requested window.
    \end{itemize}

    \subsection{Data leakage prevention}
    Data leakage (also known as target leakage) is the use of information in the model training process which would not be expected to be available at prediction time, causing the predictive scores (metrics) to overestimate the model's utility when running in a production environment. 

    The query subsystem in the feature store is designed to prevent data leakage. In the feature retrieval query, given an observation event at time \(ts_0\), the query subsystem should do:
    \begin{itemize}
        \item Only look for feature values from the past of \(ts_0\)
        \item Find the feature value from the nearest past of \(ts_0\) while considering the expected delay of source and feature data.
    \end{itemize}

    \subsection{Storage partitioning scheme}
        \subsubsection{Abstracted data structure of a materialized feature set record}
        We materialize feature data at feature set level, not feature level.  Each record of a feature set has the following fields:
        \begin{itemize}
            \item ID(s): multiple ID (index) columns
            \item Event timestamp: this is the feature value timestamp of the feature set record. I.e. in a daily aggregation Feature Set, this will be the timestamp of the end of day. It only applies to time-series based feature.
            \item Creation timestamp: this is the time that the feature set record is materialized. This is always larger than the event timestamp.  
            \item Feature(s): feature columns  
        \end{itemize}

        For a given feature set version, \(ID(s) + event\_timestamp + creation\_timestamp\) gives the uniqueness of each record.  Offline and Online store will have their own design of storing the abstracted data structure, but it should be in complete form for each record. In one implementation, offline store uses delta table to store this data structure.

        \subsubsection{Consistency between offline/online store }
        In a Feature Set version (table), for a given ID(combo), it will have multiple records over time. At any moment
        \begin{itemize}
            \item Offline store: keeps every record for each ID (combo)           
            \begin{multline}
                (event\_timestamp + \\ creation\_timestamp)
            \end{multline}
            
            \item Online store: keeps the latest record for each ID (combo), assuming TTL satisfies
            \begin{multline}
                max(tuple(event\_timestamp, \\ creation\_timestamp))
            \end{multline}
        \end{itemize}

        \begin{figure}[h]
        \centering
        \resizebox{\hsize}{!}{\includegraphics[clip=true]{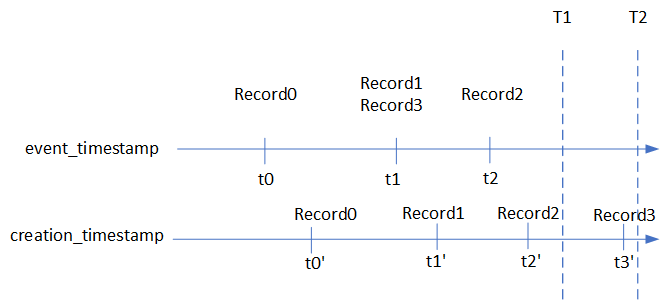}}
        \caption{
        \footnotesize
        Example records}
        \label{fig:records}
        \end{figure}

        Assuming that there is no failure of dumping the calculated Feature Set records into offline/online store in the materialization job and ignoring the latency from the moment that the Feature Set dataframe is calculated in memory to the moment that it is stored in online/offline store, then using the example of the following records as shown in Figure \ref{fig:records}.

            \(Record_0 = \{event\_timestamp:t_0, \\ creation\_timestamp:t_0’\}, (t_0’ > t_0)\) \\
            
            \(Record_1 = \{event\_timestamp:t_1, \\ creation\_timestamp:t_1’\}, (t_1’ > t_1)\) \\

            \(Record_2 = \{event\_timestamp:t_2, \\ creation\_timestamp:t_2’\}, (t_2’ > t_2)\) \\
            
            \(Record_3 = \{event\_timestamp:t_1, \\ creation\_timestamp:t_3’\}, (t_3’ >t_2’> t_1’ > t_0’)\) \\

        \noindent At \(T_1\): 
        \begin{itemize}
            \item Offline store has 3 records \({R_0, R_1, R_2}\) 
            \item Online store has 1 record \({R_2}\) 
        \end{itemize}
        
         \noindent At \(T_2\): 
         \begin{itemize}
             \item Offline store has all 4 records \({R_0, R_1, R_2, R_3}\) 
             \item Online store has 1 record \({R_2}\)
         \end{itemize}
        
        Materialization jobs may fail. However, online/offline store should reach eventual consistency with job retries (manual or auto).

        \subsubsection{Merge Feature Set records into offline/online store table}
        Each run of a materialization job (backfill or recurrent) generates a table of feature set records of a given feature set, which contains new or updated records of a given feature (event timestamp) window.  The job will merge the new table into the existing table of the feature set, with the logic in snippet \ref{alg:cap}. Offline and online store will have their own mechanism of implementing such merge logic.

        \begin{algorithm*}
        \caption{Merge Featureset logic}\label{alg:cap}
        \begin{algorithmic}
        \Require\\
        $storeType \in {offline, online}$
        \\
        \If{storeType = offline}
            \If{key(IDs + event\_timestamp + creation\_timestamp) does not exist}
               \State Insert the record
            \Else
               \State No-op
            \EndIf    
        \ElsIf{storeType = online}
            \If{key(IDs) does not exist}
               \State Insert the record
            \Else
               \If{new event\_timestamp gt existing event\_timestamp}
                   \State Override the record
               \ElsIf{new event\_timestamp = existing event\_timestamp and  new creation\_timestamp gt existing creation\_timestamp}
                   \State Override the record
               \Else
                   \State No-op
               \EndIf
            \EndIf
        \EndIf
        \end{algorithmic}
        \end{algorithm*}
        
        \subsubsection{Eventual consistency between offline/online store}
        Every time a materialization job runs (backfill or scheduled), it generates a new dataframe of the corresponding feature (event timestamp) window. If customers enable both online and offline store, that same table must be merged into both online and offline store. If the dataframe is only merged into one but not the other, it will break the eventual consistency. In eventual consistency, the time that the new table merge into offline/online completes may be different based on the design of materialization, due to reasons such as:
        \begin{itemize}
            \item Failed in one merge but not the other (and retry succeeds) 
            \item Sequence of processing the merge (e.g. offline first then online)
        \end{itemize}

        \subsubsection{Bootstrap offline to online and online to offline}
        Users may enable only one store first and later enable the other one. When they enable the second store, there is a need to bring consistency between the two stores. One option is for the user to run a backfill job that covers the feature event timestamp window from the earliest time that the first store started materialization, till now. There are two downsides: 
        \begin{itemize}
            \item Source data may not exist already for the early times.
            \item It is unnecessarily expensive, in the case that offline store is enabled first. 
        \end{itemize}
        
        So, there is a need of bootstrap the second store using data from the first store 
        \begin{itemize}
            \item Offline to online:  read the table in offline store, for each ID(s), get the record with \(\max(tuple(event\_timstamp, \\ creation\_timestamp))\), and dump to online store 
            \item Online to offline: dump everything in the online store to offline store
        \end{itemize}

    \subsection{Feature-Model lineage}
    Tracking the lineage between feature and model is useful to users. However, there are a challenges in that:

    \begin{itemize}
        \item Scalability challenge: It is not uncommon that a model can use hundreds or more features. 
        \item Cross region lineage: While the feature store is provisioned in one region, models using features from the feature store can be deployed to any other regions.
    \end{itemize}

    A dedicated subsystem in feature store should be able to track lineage to the requested scale and also provide a global view of the lineage.

\section{Related work}

There are two popular OSS solutions, Feast [\cite{feast_2023}] and Feathr [\cite{feathr-ai}].

Feast is a Python based all-client architecture feature store. Customers use its Python SDK to interact with the feature store.

Feathr is a Scala/Java based feature store derived from LinkedIn’s in-house feature store solution. Feathr is a more full-fledged OSS feature store offering however it is not managed. Most importantly, it allows defining feature transformation using both DSL and UDF. Besides that, it also has nice additions such as UX support (for managing the feature repository), and Purview integration. Feathr also claims to offer an optimized query execution.

There are also several commercial feature store solutions [\cite{databricks, google, hopsworks, hudgeon_nichol_2020, tecton}] from various vendors that exist today. 

\section{Future directions}\label{sec:future}
With the recent advancements of LLMs and vector databases, we see a need to enhance feature stores to support non time series representation which can support range queries. Such range queries are crucial to support vector search. We would also like to invest even more in our infra services to improve managed Spark compute experiences thereby providing a delightful experience to customers. Supporting federated learning ecosystems [\cite{matschinske2021featurecloud}] is another exploration area.

\section{Conclusion}\label{sec:conclusion}
In this paper, we have discussed the architecture and system design of the managed feature store. A managed feature store solution requires several subsystems to tackle the challenges including offline/online data consistency, data leakage prevention, cross-region/platform asset sharing, and more. In future we plan to explore even more optimizations to provide an even better cost-effective offering for our customers. While some of the components are aspirational and not implemented in AzureML Feature Store yet, the overall design represent our strategic ideation and offer a glimpse into the prospective capabilities of the product.

\section{Acknowledgements}\label{sec:ack}
We would like to thank all our customers for their valuable feedback. We are also grateful to our AzureML [\cite{azure-machine-learning}] colleagues and partners who supported us with our vision and were also part of our journey in building such a novel system.

\bibliographystyle{acm} 
\bibliography{refs} 

\begin{thebibliography}{10}

\bibitem{delta-lake}
{\sc Armbrust, M., et~al.}
\newblock Delta lake: High-performance acid table storage over cloud object
  stores.
\newblock {\em Proc. VLDB Endow. 13}, 12 (aug 2020), 3411–3424.

\bibitem{databricks}
{\sc Databricks}.
\newblock Databricks feature store.

\bibitem{feast_2023}
{\sc Feast}.
\newblock Feature store for machine learning, Jan 2023.

\bibitem{feathr-ai}
{\sc Feathr}.
\newblock Feathr-ai/feathr: Feathr – a scalable, unified data and ai
  engineering platform for enterprise.

\bibitem{google}
{\sc Google}.
\newblock Vertex ai documentation.

\bibitem{hopsworks}
{\sc Hopsworks}.
\newblock The python-centric feature store.

\bibitem{hudgeon_nichol_2020}
{\sc Hudgeon, D., and Nichol, R.}
\newblock Machine learning for business: Using amazon sagemaker and jupyter,
  2020.

\bibitem{matschinske2021featurecloud}
{\sc Matschinske, J., et~al.}
\newblock The featurecloud ai store for federated learning in biomedicine and
  beyond, 2021.

\bibitem{azure-machine-learning}
{\sc Microsoft}.
\newblock Azure machine learning - ml as a service.

\bibitem{mj_2022}
{\sc Mj, J.~K.}
\newblock {\em Feature Store for Machine Learning: Making features sharable and
  reproducible}.
\newblock Packt Publishing, 2022.

\bibitem{managing-ml-pipeline}
{\sc Orr, L., et~al.}
\newblock Managing ml pipelines: Feature stores and the coming wave of
  embedding ecosystems.
\newblock {\em Proc. VLDB Endow. 14}, 12 (jul 2021), 3178–3181.

\bibitem{redis_2023}
{\sc Redis}.
\newblock Vector database and vector similarity search, Apr 2023.

\bibitem{tecton}
{\sc Tecton}.
\newblock Tecton documentation: Tecton.

\end{thebibliography}

\end{document}